\documentclass[letterpaper, 10 pt, conference]{ieeeconf}  % Comment this line out
\IEEEoverridecommandlockouts                              % This command is only
\usepackage{enumerate}
\usepackage{graphicx}
\usepackage{ams}
\usepackage{booktabs}
\usepackage{setspace}
\usepackage[dvips]{color}
\usepackage{epsf}
\usepackage{times}
\usepackage{epsfig}
\usepackage{graphicx}
\usepackage{url}
\usepackage{amsmath}
\usepackage{amssymb}
\usepackage{amsxtra}
\usepackage{algorithmic}
\usepackage{algorithm}
\usepackage{cite}
\usepackage{flushend}

\newtheorem{assumption}[subsubsection]{\bf{Assumption}}
\newtheorem{subthm}[subsubsection]{\bf{Theorem}}

\newtheorem{subcoro}[subsubsection]{\bf{Corollary}}

\newtheorem{sublem}[subsubsection]{\bf{Lemma}}

\newtheorem{subpropo}[subsubsection]{\bf{Proposition}}
\newtheorem{definition}[subsection]{\bf{Definition}}

\newtheorem{subrem}[subsubsection]{Remark}

\def\ind{{\mathchoice {\rm 1\mskip-4mu l} {\rm 1\mskip-4mu l}
{\rm 1\mskip-4.5mu l} {\rm 1\mskip-5mu l}}}

\title{\Large
Heterogeneous Learning in Zero-Sum  Stochastic Games with Incomplete Information \vspace{-3mm}}

\author{Quanyan Zhu$^\dag$, Hamidou Tembine$^\ddag$ and Tamer Ba\c{s}ar$^\dag$ \thanks{This work was supported in part by a  grant from AFOSR.}
\thanks{$^\dag$Q. Zhu and T. Ba\c{s}ar are with Dept. ECE and CSL, University of Illinois, 1308 West Main, Urbana, IL, 61801, USA.        {\tt\small \{zhu31, basar1\}@illinois.edu}}
\thanks{$^\ddag$H. Tembine is with Sup\'elec,  3 rue Joliot-Curie 91192 Gif-sur-Yvette cedex, France
        {\tt\small tembine@ieee.org}}%
}

\begin{document}
 %\singlespacing

\maketitle
%\thispagestyle{empty}
%\pagestyle{empty}

%\title{Heterogeneous learning in zero-sum stochastic games}
%

%TB Replaced the abstract with what I had sent for the invited session
\vspace{-5mm}\begin{abstract}
Learning algorithms are essential for the applications of game theory in a networking environment. In dynamic and decentralized settings where the traffic, topology and channel states may vary over time and the communication between agents is impractical, it is important to formulate and study games of incomplete information and fully distributed learning algorithms which for each agent requires a minimal amount of information regarding the remaining agents. In this paper, we address this major challenge and introduce heterogeneous learning schemes in which each agent adopts a distinct learning pattern in the context of games with incomplete information. We use stochastic approximation techniques to show that the heterogeneous learning schemes can be studied in terms of their deterministic ordinary differential equation (ODE) counterparts. Depending on the learning rates of the players, these ODEs could be different from the standard replicator dynamics, (myopic) best response (BR) dynamics, logit dynamics, and fictitious play dynamics. We apply the results to a class of security games in which the attacker and the defender adopt different learning schemes due to differences in their rationality levels and the information they acquire.
\end{abstract}

%%%tembine: heterogeneous or heterogenous? what is the difference if any?
 \section{Introduction}
 Distributed iterative schemes play an important role in the computation of equilibria and the estimation of payoffs under incomplete information \cite{basar99}. This paper studies
a  two-person zero-sum stochastic game  with an arbitrary number of states and  a finite number of actions for each player.
 When each player has a complete knowledge of its payoff function and has past access to  past actions of the others, then there is an arsenal of tools such as  fictitious play algorithms, best response dynamics, and gradient-based algorithms, that can be used to arrive at the equilibrium of the game. However, it is well known that these algorithms may fail to converge even under the perfect observation of actions and payoffs \cite{fail3,young,leslie,benaim2}.
A new learning challenge hence
arises when a player  does not know its own payoff function
 and/or has no information about the past actions of the other players.
 In this case, the player needs to interact with the environment to find out its expected payoff and its optimal strategy.
 %The player can learn  both expected payoff and the optimal strategy by using some iterative techniques based only own-experiences.

%We address one of the major challenge in dynamic interaction,
In practical applications, we are often in search of distributed learning algorithms
that require a minimal amount of information and a minimal
amount of resources. It is then natural to ask whether there exists a learning scheme that demands less information and less
memory within a dynamically evolving environment, and
leads to an efficient, stable and fair outcome.
In this paper, we address this challenge by proposing a class of heterogeneous learning algorithms in a scenario where  the players do not know their own payoff functions. At each time $t$, each player chooses an action and receives a numerical value for its payoff or {\it perceived payoff} as an outcome of the instantaneous game.
In contrast to fictitious play and best response dynamics which require the knowledge of the history of actions played by the other players, our learning algorithm relaxes this assumption. Indeed, it is often implausible and impractical in applications to assume  the capability of observations of the actions of the other players. Furthermore, we assume that the state space of the game and its transition law between the states are  unknown to the players. In addition, the players also do not have the knowledge of  the {\it action spaces} of the others. %The state space of the stochastic game is also unknown to the players. %A player does not know the action space of the others.
The question we will address is  how much the players can expect to learn under such circumstances?

We propose different coupled (or combined) and fully distributed learning schemes that enable learning optimal strategies and concurrently estimating the optimal payoffs. In contrast to the standard reinforcement learning algorithms which focus only on either strategy or payoff reinforcement for the equilibrium learning, the algorithm that couples the payoff-reinforcement learning together with strategy-reinforcement learning enables an immediate prediction and updates the strategies by updated estimations based on recent experiences.
Our learning algorithms also offer the degrees of freedom to model different levels of rationality and  learning rates of the players.
The  ordinary differential equations (ODEs)  associated with the stochastic learning algorithms differ from the standard replicator dynamics, best response dynamics and fictitious play dynamics. Particular connections to {\it logit} dynamics and {\it imitative logit dynamics} are also established. Using basic stochastic approximation
techniques from \cite{borkar97,kushner78,benaim2,leslie} and under suitable assumptions on the  learning rates, we show  their convergence to a new class of game dynamics and asymptotic properties of different learning algorithms  within a class of zero-sum stochastic games.
%\subsection{Structure}

The paper is structured as follows. In next section, we present the zero-sum stochastic game model and provide an overview of the basic properties of  reinforcement learning algorithms. Section \ref{main} presents our main results on heterogeneous learning algorithms. In Section IV, we apply the learning algorithms to study security games and provide numerical results. Section \ref{con} concludes the paper and discusses future work. %We summarize some of the notation used  in the paper in Table I for the convenience of the reader.
%
%\begin{table}[t]
%\caption{Glossary}
%\begin{center}
%\begin{tabular}{ll@{}c@{}lc}
%\toprule
%  Symbol & Meaning  \\
%\midrule
%%  \hline
%%  \hline
%  $a_{j,t}$  & Action of player $j$ (P$j$) at time $t$ \\
%  $ \mathbf{f}_t$ & Mixed strategy of P$1$ at $t$ \\
%   $\mathbf{g}_{t}$ & Mixed strategy of P$2$ at $t$ \\
%  $U_{j,t}$  & Perceived payoff by $j$ at $t$ \\
%  $\hat{u}_{j,t}$ & Estimated payoff vector of $j$ at $t$ \\
%   $\mathbb{U}_j$ & Mixed extension of the payoff function  $U_{j}$.\\
%  $u_j(\mathbf{f},\mathbf{g})$ & Expected payoff $\mathbb{E}_{s,\mathbf{f},\mathbf{g}}{U}_j(s,a_1,a_2)$\\
%  $\tilde{\beta}_{j,\epsilon}(\hat{u}_{j,t})$  & Boltzmann-Gibbs strategy   \\
%    ${\beta}_{1,\epsilon}(\mathbf{g}), {\beta}_{2,\epsilon}(\mathbf{f})$  & Smooth best responses   \\
%  $\sigma_j(\hat{u}_{j,t})$  & Imitative Boltzmann-Gibbs strategy   \\
%  $\mu_{j,t}$ & Payoff learning rates of P$j$ at $t$\\
%  $\lambda_{j,t}$ & Strategy learning rates of P$j$ at $t$\\
%  $e_{a_i}$ & The unit vector with $1$ at the position of $a_i$ \\ & and $0$ otherwise\\ \bottomrule
%\end{tabular}
%\end{center}
%\label{notation}
%\end{table}%

\section{Game Model and Learning Algorithms} \label{secmodel}
In this section, we formulate a two-person zero-sum stochastic game model $\Xi=\langle \mathcal{S}, \mathcal{A}_1, \mathcal{A}_2, \{U(s,.)\}_{s\in \mathcal{S}} \rangle$ where  $\mathcal{A}_1,\mathcal{A}_2$ are the finite sets of  actions available to players P1 and P2, respectively, and $\mathcal{S}$ is the set of possible states. We assume that the state space $\mathcal{S}$ and the
%TB probability transition
probability distribution
on the states are both unknown to the players. A state $s\in\mathcal{S}$ is an independent and
identically distributed random variable defined on the set $\mathcal{S}$. % (identically of the chosen actions).
We assume the action spaces are the same in each state.
The zero-sum game is characterized by a single utility function $U: \mathcal{S}\times\mathcal{A}_1\times\mathcal{A}_2\rightarrow \mathbb{R}$.  P1 collects a payoff $U_1(s, a_1, a_2)=U(s, a_1, a_2)$ when he chooses $a_1\in\mathcal{A}_1$ and P2 uses $a_2\in\mathcal{A}_2$ at state $s\in\mathcal{S}$, and for the same choices P2 collects a payoff
of $U_2(s, a_1, a_2)=c-U(s, a_1, a_2);$ equivalently,  $U(s, a_1, a_2)-c$ is cost to P2, where $c$ is a constant. In terms of the single utility function $U$,  P1 is the maximizer and P2 is the minimizer, and both players are interested in the performance at steady state using mixed strategies, as to be made clear shortly.
% P1 decides on certain strategies to maximize his payoff while P2 has his conflicting goal against P1. P2's instantaneous payoff is given by $U_2$ such that $U_1(s,.)+U_2(s,.)=c$, for some constant $c\in\mathbb{R}$.
The preceding game model can be viewed as a special class of stochastic games in which the state transitions are independent of the player actions as well as the current state. Note that what we have here is a constant-sum game, where the constant is $c$. In the analysis of its equilibrium, we can let $c=0$ without any loss of generality, and hence view it as a zero-sum game.

We have slotted time, $t\in\{0, 1, \ldots\}$, when players pick their mixed strategies as functions of what has transpired in the past, to the extent the information available to them allows. Toward this end, we let   $f_{t}(a_1)$ and $g_{t}(a_2)$ denote the probabilities of P$1$ choosing $a_1\in\mathcal{A}_1$ and P2 choosing $a_2\in\mathcal{A}_2$,  respectively,  at time $t$, and let $\mathbf{f}_t=[f_t(a_1)]_{a_1\in\mathcal{A}_1} $ and $\mathbf{g}_t=[g_t(a_2)]_{a_2\in\mathcal{A}_2}$ be the mixed strategies of P1 and P2 respectively (at time $t$), where more precisely
{\begin{eqnarray}
\mathbf{f}_t\in \mathcal{F}:=\left\{ \mathbf{f}:\ f(a_1)\in [0, 1] , \sum_{a_1\in\mathcal{A}_1}f( a_1)=1\right\};\\
 \mathbf{g}_t\in \mathcal{G}:=\left\{ \mathbf{g}:\ g( a_2)\in [0, 1], \sum_{a_2\in\mathcal{A}_2}g(a_2)=1\right\}.
 \end{eqnarray}}
In particular, we define $e_{a_1}, e_{a_2},$ with $a_1\in\mathcal{A}_1, a_2\in\mathcal{A}_2,$ as unit vectors of sizes $|\mathcal{A}_1|$ and $|\mathcal{A}_2|$ , respectively, whose entry that corresponds to $a_1$ or $a_2$ is 1 while others are zeros.
 We assume that the mixed strategies of the players are independent of the current state $s.$
For any given pair of mixed strategies, $(\mathbf{f}\in \mathcal{F}, \mathbf{g}\in \mathcal{G}),$ and for a fixed $s\in S$, we define the expected utility (as expected payoff to P1 and expected cost to P2) as
$\mathbb{U}(s, \mathbf{f} ,\mathbf{g}):= \mathbb{E}_{\mathbf{f} ,\mathbf{g}} U(s, a_1, a_2),$
where $\mathbb{E}_{\mathbf{f} ,\mathbf{g}} $ denotes expectation of $U$ over the action sets of the players under the given mixed strategies. A further expectation of this quantity over the states $s$, denoted $\mathbb{E}_s$, yields the performance index of the \textit{expected game}. We now define the equilibrium concept of interest for this game, that is the saddle-point equilibrium:
\begin{definition}[Saddle Point]
A  strategy pair $(\mathbf{f}^*, \mathbf{g}^*)$ constitutes a {\em saddle point} for the expected game if and only if  $\forall \mathbf{f}\in\mathcal{F}$ and $\mathbf{g}\in\mathcal{G}$,
\begin{equation}
\mathbb{E}_s\mathbb{U}(s, \mathbf{f} ,\mathbf{g}^*)\leq \mathbb{E}_s \mathbb{U}(s, \mathbf{f}^* ,\mathbf{g}^*)\leq \mathbb{E}_s\mathbb{U}(s, \mathbf{f}^* ,\mathbf{g}).
\end{equation}
\end{definition}

This now being a finite zero-sum game (or constant sum game, if $c\neq0$), the existence of a saddle point is guaranteed by the minimax theorem.
%since the mapping $\mathcal{ME}: (\mathbf{f},\mathbf{g})\longmapsto\mathbb{E}_s U(s, \mathbf{f} ,\mathbf{g})$ is jointly continuous, concave in $\mathbf{f}$, convex in $\mathbf{g}$ and the spaces $\mathcal{F},$ $\mathcal{G}$ are non-empty, convex and compact.
%%%tembine: zero-sum repeated games with random parameters (independent)  or
%%%%%%%%%%: or zero-sum stochastic games with indpendent state-transition and strategies are state-indpendent

We now consider this game played over the discrete-time horizon, with the players generating mixed strategies, say  $( \mathbf{f}_t, \mathbf{g}_t)$ at every time point $t$. These strategies will be generated (recursively updated) according to some rule, which uses the information available to the players.
As indicated before, the players do not know the functional form of $U$, that is they do not know the entries of the underlying matrix, but at each time $t$ they observe the value $U(s, a_{1,t}, a_{2,t})$, where the actions are realized under $( \mathbf{f}_t, \mathbf{g}_t)$, and they recall their own past actions. With this information, P1 and P2 generate, respectively,  $\mathbf{f}_{t+1}$ and $ \mathbf{g}_{t+1}$.  The precise way of doing this is determined by the algorithm picked, and there will be several such algorithms as will be discussed shortly. For each one, our goal is to show that the sequences thus generated converge to the pair of mixed saddle-point strategies, that is
$ \lim_{t\to\infty} \mathbf{f}_t  = \mathbf{f}^*,\  \lim_{t\to\infty} \mathbf{g}_t  = \mathbf{g}^*,$
where the \textit{limit} will be given a precise meaning later.

%Both players choose a mixed strategy pair   ($\mathbf{f},\mathbf{g}$)  to optimize its payoff in the long-run game. % $v(s, \mathbf{f}, \mathbf{g} )$ by choosing  a mixed strategy pair ($\mathbf{f},\mathbf{g}).
%Since a player does not observe the past actions of the other player, we consider strategies used by both players to be only dependent on their current perceived payoffs and their own actions that  have been used. We define a {\it stationary saddle point} of the expected game  in the following definition.
%With some abusive notation, we use the mixed extension payoff under the same name as the payoff defined on action profiles. The pure strategies (Dirac distribution over actions) and the actions are both taken in argument.
%Likewise, for $\epsilon\geq 0,$ we can define an $\epsilon-$ {\it stationary saddle point} as follows.
%\begin{definition}[$\epsilon-$Saddle-Point]
%A  strategy pair $(\mathbf{f}^*, \mathbf{g}^*)$ is a mixed  {\em $\epsilon-$ stationary saddle-point} for the expected game if and only if
%\begin{equation}
%\mathbb{E}_sU(s, \mathbf{f} ,\mathbf{g}^*)-\epsilon \leq \mathbb{E}_s U(s, \mathbf{f}^* ,\mathbf{g}^*)\leq \mathbb{E}_sU(s, \mathbf{f}^* ,\mathbf{g})+\epsilon,
%\end{equation}
 %$\forall \mathbf{f}\in\mathcal{F}$ and $\mathbf{g}\in\mathcal{G}$.
%\end{definition}
%TB LEFT HERE
\subsection*{A. Learning Schemes}
To achieve the saddle-point solution, we suggest the following reinforcement learning mechanism for homogeneous
learners. We use the abbreviation ``RL"  
for ``reinforcement learning" and  ``C"   for ``combined", suggesting that the algorithm involves
learning the expected utility as well as the strategies. %$\bullet$
%%%%
%describe the general form of the algorithms here.
We consider  combined fully distributed,  payoff and strategy reinforcement learning (CODIPAS-RL) in the form:
$$ \ \left\{\begin{array}{ccc}
\mathbf{f}_{t+1}&=& \mathbf{f}_{t}+\Pi_{11}(\lambda_{1,t}, a_{1,t},U_{1,t}, \hat{\mathbf{u}}_{1,t}, \mathbf{f}_{t} )\\
\hat{\mathbf{u}}_{1,t+1}&=& \hat{\mathbf{u}}_{1,t}+\Pi_{12}(\mu_{1,t}, a_{1,t}, U_{1,t},\mathbf{f}_{t}, \hat{\mathbf{u}}_{1,t}) \ \\
\mathbf{g}_{t+1}&=& \mathbf{g}_{t}+\Pi_{21}(\lambda_{2,t}, a_{2,t},U_{2,t}, \hat{\mathbf{u}}_{2,t}, \mathbf{g}_{t} )\\
\hat{\mathbf{u}}_{2,t+1}&=& \hat{\mathbf{u}}_{2,t}+\Pi_{22}(\mu_{2,t}, a_{2,t}, U_{2,t},\mathbf{g}_{t}, \hat{\mathbf{u}}_{2,t}) \ \\
&&  t\geq 0, a_{i,t}\in\mathcal{A}_i, i\in\{1,2\},
\end{array}
\right.
$$
where $\Pi_{i1}, \Pi_{i2},  i\in\{1,2\},$ are properly chosen functions. %such that  $\forall t,\ \mathbf{f}_t\in\mathcal{F},\ \mathbf{g}_t \in\mathcal{G}.$
The parameters $\lambda_{i,t}, \mu_{i,t}$ are learning rates indicating players'  capabilities of information retrieval and update. The vectors $\mathbf{f}_t \in\mathcal{F}, \mathbf{g}_t\in\mathcal{G}$ are mixed strategies of the players at time $t$. $\hat{\mathbf{u}}_{i,t}, i\in\{1,2\},$ are estimated average payoffs updated at each iteration $t$, and $U_{i, t}, i\in\{1,2\},$ are the perceived payoffs received by players at time $t$.
%%%%%%%%%%%%%%%%%%%%%%%%%%%ù

We identify below five different special cases of this general class of learning algorithms, each one important in its own right.

\subsubsection{CRL0}
The first COmbined fully DIstributed PAyoff and Strategy Reinforcement Learning (CODIPAS-RL) algorithm is CRL0  given in (\ref{CRL0}) below, which captures the procedure in \cite{fail3} for both payoffs and strategies. At every time step $t$,  P1 and P2 each chooses an action according to their estimations and their mixed strategy vectors $\mathbf{f}_t$ and $\mathbf{g}_t$, respectively. Based on the joint action, each player perceives his instantaneous payoff $U_{i,t}$, $i\in\{1, 2\}$, and updates his strategy vectors. The strategy and utility updates  are not coupled and do not involve optimal choices of the players. The players make updates by taking a weighted average of the current observed payoff and the quantities from the previous iteration.  The indicator function $\ind_{\{ a_{i,t}\}}$ is a unit vector of appropriate dimension with one of its components corresponding to the action chosen at time $t$, $a_{i,t}$, being $1$ and the others being zeros. The step size parameters $\lambda_{i,t}$ need to be small enough such that $\lambda_{i,t} U_{i,t} <1$ for all $t$. 
\begin{equation}\label{CRL0}
\left\{\begin{array}{lll}
\mathbf{f}_{t+1}&=& \mathbf{f}_{t}+\lambda_{1,t} {U}_{1,t}\cdot \left(\ind_{\{ a_{1,t}=a_1\}}-\mathbf{f}_{t} \right)\\
\hat{\mathbf{u}}_{1,t+1}&=&\hat{\mathbf{u}}_{1,t}+\mu_{1,t}\ind_{\{ a_{1,t}=a_1\}}\left( U_{1,t}-\hat{\mathbf{u}}_{1,t}\right) ,
\ a_1\in\mathcal{A}_1\\
\mathbf{g}_{t+1}&=& \mathbf{g}_{t}+\lambda_{2,t} {U}_{2,t}\cdot\left(\ind_{\{ a_{2,t}=a_2\}}-\mathbf{g}_{t} \right)\\
\hat{\mathbf{u}}_{2,t+1}&=&\hat{\mathbf{u}}_{2,t}+\mu_{2,t}\ind_{\{ a_{2,t}=a_2\}}\left( U_{2,t}-\hat{\mathbf{u}}_{2,t}\right) ,
\ a_2\in\mathcal{A}_2\\
\end{array}
\right.
\end{equation}
\subsubsection{CRL1}
Algorithm CRL1 given in (\ref{CRL1}) below is another combined algorithm that learns the average utility and the mixed strategies concurrently. This is a Boltzmann-Gibbs based CODIPAS-RL. In a similar fashion as in CRL0, P1 and P2 select their actions based on their current strategy distributions. However, the updates on the strategies and the average payoff follow  reinforcement learning and $\lambda_{i,t}$ and $\mu_{i,t}$ are the learning rates for the payoffs and the strategies respectively, satisfying Assumption \ref{RateAssumption} and $\frac{\lambda_{i,t}}{\mu_{i,t}}\rightarrow0, i\in\{1, 2\}$.
\begin{equation}\label{CRL1}
\left\{\begin{array}{lll}
\mathbf{f}_{t+1}&=&(1-\lambda_{1,t})\mathbf{f}_{t}+\lambda_{1,t} \tilde{\beta}_{1,\epsilon}(\hat{\mathbf{u}}_{1,t})\\
\hat{\mathbf{u}}_{1,t+1}&=&\hat{\mathbf{u}}_{1,t}+\frac{\mu_{1,t}}{f_{t}(a_1)}\ind_{\{ a_{1,t}=a_1\}}\left( U_{1,t}-\hat{\mathbf{u}}_{1,t}\right), \ a_1\in\mathcal{A}_1\\
\mathbf{g}_{t+1}&=&(1-\lambda_{2,t})\mathbf{g}_{t}+\lambda_{2,t} \tilde{\beta}_{2,\epsilon}(\hat{\mathbf{u}}_{2,t})\\
\hat{\mathbf{u}}_{2,t+1}&=&\hat{\mathbf{u}}_{2,t}+\frac{\mu_{2,t}}{g_{t}(a_2)}\ind_{\{ a_{2,t}=a_2\}}\left( U_{2,t}-\hat{\mathbf{u}}_{2,t}\right), \ a_2\in\mathcal{A}_2
\end{array}
\right.
\end{equation}
where $\tilde{\beta}_{i,\epsilon}: \mathbb{R}^{|\mathcal{A}_i|}\rightarrow\mathbb{R}^{|\mathcal{A}_i|}, i\in\{1,2\},$ is the  Boltzmann-Gibbs strategy or the soft-max function parameterized by $\epsilon\geq 0$,  which takes in the average payoff vector and produces a vector that assigns more weight to the maximum component. The weight assigned to a particular action $a_i\in \mathcal{A}_i, i\in\{1, 2\}$ is given by
\begin{equation}
\tilde{\beta}_{i,\epsilon}(\hat{\mathbf{u}}_{i,t})(a_i)=
\frac{e^{\frac{1}{\epsilon}\hat{u}_{i,t}(a_i)}}{\sum_{a_i'}e^{\frac{1}{\epsilon}
\hat{u}_{i,t}(a'_i)}}, a_i\in\mathcal{A}_i, i\in\{1,2\}.
\end{equation}
It is clear that when $\epsilon$ is high, the output of the $\tilde{\beta}_{i,\epsilon}$ function does not distinguish among the actions and assign equal weights to them; when $\epsilon$ approaches zero, $\tilde{\beta}_{i,\epsilon}$ function bears more resemblance with the maximum function, assigning $1$ to the action yielding the maximum average payoff but zeros to the other actions \cite{Shamma}.

\subsubsection{CRL2}
The procedure for the CODIPAS-RL algorithm  CRL2 is similar to CRL1  but only differs in the use of soft-max function. In place of the  Boltzmann-Gibbs strategy, we adopt imitative Boltzmann-Gibbs strategy which is weighted by the current strategy vector \cite{hofbauer98}, and is given by $\sigma_i : \mathbb{R}^{|\mathcal{A}_i|}\times \mathbb{R}^{|\mathcal{A}_i|}\rightarrow\mathbb{R}^{|\mathcal{A}_i|}, i\in\{1,2\}$. The component-wise mapping for P1 is expressed by
\begin{equation}
\sigma_1(\mathbf{f}_{t},\hat{\mathbf{u}}_{1,t})(a_1)=
\frac{f_{t}(a_1)e^{\frac{1}{\epsilon}\hat{u}_{1,t}(a_1)}}{\sum_{a_1'\in\mathcal{A}_1}f_{t}(a'_1)e^{\frac{1}{\epsilon}
\hat{u}_{1,t}(a'_1)}}.
\end{equation}
Likewise, for P2, we have
\begin{equation}
\sigma_2(\mathbf{g}_{t},\hat{\mathbf{u}}_{2,t})(a_2)=
\frac{g_{t}(a_2)e^{\frac{1}{\epsilon}\hat{u}_{2,t}(a_2)}}{\sum_{a_2'\in\mathcal{A}_2}g_{t}(a'_2)
e^{\frac{1}{\epsilon}\hat{u}_{2,t}(a'_2)}}.
\end{equation}
Collecting all this, the CRL2 algorithm is then as given below:
%$\bullet$
 %Imitation
\begin{equation}\label{CRL2}
\left\{\begin{array}{lll}
\mathbf{f}_{t+1}&=&(1-\lambda_{1,t})\mathbf{f}_{t}+\lambda_{1,t} \sigma_{1}(\mathbf{f}_{t},\hat{\mathbf{u}}_{1,t})\\
\hat{\mathbf{u}}_{1,t+1}&=&\hat{\mathbf{u}}_{1,t}+\frac{\mu_{1,t}}{f_{t}(a_1)}\ind_{\{ a_{1,t}=a_1\}}\left( U_{1,t}-\hat{\mathbf{u}}_{1,t}\right) \ \\
\mathbf{g}_{t+1}&=&(1-\lambda_{2,t})\mathbf{g}_{t}+\lambda_{2,t} \sigma_{2}(\mathbf{g}_{t},\hat{\mathbf{u}}_{2,t})\\
\hat{\mathbf{u}}_{2,t+1}&=&\hat{\mathbf{u}}_{2,t}+\frac{\mu_{2,t}}{g_{t}(a_2)}\ind_{\{ a_{2,t}=a_2\}}\left( U_{2,t}-\hat{\mathbf{u}}_{2,t}\right) \
\end{array}
\right.
\end{equation}
\subsubsection{RL2}
%%tembine: for your info: see Borger \& Sarin (1993), Arthur (1993), Sastry et al. (1994),
The learning algorithm (\ref{RL2}) updates  strategies simultaneously \cite{fail3,arthur}.
\begin{equation}\label{RL2}
\left\{\begin{array}{lll}
\mathbf{f}_{t+1}&=& \mathbf{f}_{t}+\lambda_{1,t} U_{1,t}\cdot \left( \ind_{\{ a_{1,t}=a_1\}}-\mathbf{f}_{t}\right) \ \\
 \mathbf{g}_{t+1}&=& \mathbf{g}_{t}+\lambda_{2,t} U_{2,t}\cdot \left( \ind_{\{ a_{2,t}=a_2\}}-\mathbf{g}_{t}\right)
\end{array}
\right.
\end{equation}
\subsubsection{RL3}
In RL3, we normalize RL2 by some constant $n$ and $C$. This algorithm has
appeared in \cite{arthur} and is summarized below in (\ref{RL3}):
%$\bullet$
%tembine: for your info:  %Arthur (1993), Sastry et al. (1994), Weibull (1995)
\begin{equation}\label{RL3}
 \left\{\begin{array}{lll}
\mathbf{f}_{t+1}&=&\frac{C(n+1)}{nC+U_{1,t}} \left[\mathbf{f}_{t}+U_{1,t}\ind_{\{a_{1,t}=a_1\}}\right]\\
 \mathbf{g}_{t+1}&=&\frac{C(n+1)}{nC+U_{2,t}} \left[\mathbf{g}_{t}+U_{2,t}\ind_{\{a_{2,t}=a_2\}}\right]
\end{array}
\right.
\end{equation}

The following assumption on learning rates is adopted for  all the above listed learning schemes.%, we now have the following conditions on the learning rates $\lambda_{i,t}$ and $\mu_{i,t}$:
\begin{assumption}\label{RateAssumption}
The learning rates $\lambda_{i,t}, \mu_{i,t} $,  $i\in\{1,2\}$, satisfy the following conditions:
{\begin{equation}
\lambda_{i,t}\geq 0,\ \sum_{t\geq 1}\lambda_{i,t}=+\infty,\ \sum_{t\geq 1}\lambda_{i,t}^2<+\infty,  i\in\{1,2\}
\end{equation}
\begin{equation}
\mu_{i,t}\geq 0,\ \sum_{t\geq 1}\mu_{i,t}=+\infty,\ \sum_{t\geq 1}\mu_{i,t}^2<+\infty,  i\in\{1,2\}
\end{equation}}
\end{assumption}
The learning rate which perhaps has the simplest  form that satisfies the conditions of Assumption \ref{RateAssumption} is the harmonic sequence, i.e.,
$\textrm{(R1)}\  \mu_{i,t} =\frac{1}{t+1}.$
To study learning on different time scales, we need to consider other learning rates. Typical learning rates are
$\textrm{(R2)}\ \mu_{i,t}=\frac{1}{(t+1)\log(t+1)},$
$\textrm{(R3)} \ \mu_{i,t}=\frac{1}{\sqrt{t+1}\log^2(t+1)},$
 $\textrm{(R4)}\ \mu_{i,t}=\frac{1}{(t+c')^{\rho_i}},\ \frac{1}{2}<\rho_i\leq 1,\ c'>0.$
It is clear that the learning rate (R1) is faster than (R2) and (R3). In addition, by scaling $\rho_i$ in (R4), we can obtain learning rates on different time scales.
\subsection{Basic properties}\label{basic}
\subsubsection{Properties of  RL2, RL3 and CRL0 }
The algorithm RL2 has been studied by Borgers and Sarin in \cite{fail3}.  The algorithm RL3 is a normalized version of RL2. This version has been studied by Arthur in \cite{arthur}. These authors have shown that  RL2 goes to a pseudo-trajectory of the replicator dynamics when the learning rate $\lambda_{i,t}$ goes to zero. Similarly the reinforcement learning RL3 goes to a trajectory of  an adjusted version of the replicator equation.

The learning algorithm CRL0 is obtained by combining these strategy reinforcement learnings with a payoff reinforcement learning (Q-learning). The Q-learning is known to be convergent to the expected payoffs if all the actions are sufficiently used and the learning parameters satisfy the standard conditions. The combination of these two approaches gives a new learning algorithm called {\it combined fully distributed payoff and strategy reinforcement learning} (CODIPAS-RL). With this new algorithm, the players will be able to learn both expected payoffs and the associated optimal strategies i.e., if $(\mathbf{f}_t, \hat{u}_{1,t},\mathbf{g}_t, \hat{u}_{2,t})\longrightarrow (\mathbf{f}^*, \hat{u}_{1}^*,\mathbf{g}^*, \hat{u}_{2}^*)$, then $(\mathbf{f}^*,\mathbf{g}^*)$ is a saddle point of the expected game and $\mathbb{E}_s \mathbb{U}(s,\mathbf{f}^*,\mathbf{g}^*)=\hat{u}_{1}^* = c-\hat{u}^*_2.$ Moreover, the strategies are generated by the replicator equation:
\begin{eqnarray}
\nonumber \dot{f}_t(a_1)&=& {f}_t(a_1)[ {u}_1(e_{a_1},\mathbf{g}_t)-\sum_{a'_1\in\mathcal{A}_1} {u}_2(e_{a'_1}, \mathbf{g}_t) f_t(a'_1) ]\\
\nonumber \dot{g}_t(a_2)&=& {g}_t(a_2)[ {u}_2(\mathbf{f}_t,e_{a_2})-\sum_{a'_2\in\mathcal{A}_2} {u}_2(\mathbf{f}_t,e_{a'_2}) g_t(a'_2) ]
\end{eqnarray}
where  $u_1(\mathbf{f}^*,\mathbf{g}^*)=\mathbb{E}_s \mathbb{U}(s,\mathbf{f}^*,\mathbf{g}^*)$ and $u_2(.)=c-u_1(.).$

%We first recall  the folk theorem %(evolutionary version)
%under the replicator dynamics.
% \begin{subthm}[\cite{hofbauer98}] The replicator dynamics of the   expected two-person game has the following properties.
%\begin{enumerate}[(P1)]
%\item Every saddle point of the expected game is a rest  point.
%\item Every strict saddle point  of the expected game is asymptotically stable \cite{hofbauer98}.
%\item Every stable rest point is a saddle point of the expected game.
%\item If an interior orbit converges, its limit is a saddle point of the expected game.
%\end{enumerate}
%\end{subthm}
%We refer the reader to \cite{hofbauer09} for more recent analysis of the stochastic replicator dynamics. In the two-person case, there are many other interesting properties. %Let $\mathcal{A}_1=\mathcal{A}_2.$ If the time average $\frac{1}{T}\int_0^T {f}_{t}(a_1)\ dt$ of an interior orbit converges, the limit is a Nash equilibrium.
%% The following result follows from \cite{hofbauer09,hofbauer98}.
%\begin{subthm}[\cite{hofbauer09,hofbauer98}]
%For the learning schemes RL2, RL3, CRL0,  the  time averages $$\left(\frac{1}{T}\int_0^T{f}_{t}(a_1)\ dt,\ \frac{1}{T}\int_0^T{g}_{t}(a_2)\ dt\right)$$ of an interior orbit converge to the set of saddle points.
%\end{subthm}
%In addition, the limit set of $P_{a_1a_2}^T:=\frac{1}{T}\int_0^T
%{f}_{t}(a_1) {f}_{t}(a_2) \ dt$ leads to  coarse correlated equilibria, \cite{young}.

A major inconvenience with  CODIPAS-RL, CRL0,   RL2 and   RL3 is that the rest points (equilibrium states) of the corresponding  ODEs are not necessarily equilibria of the expected game. For example, all the faces of the  simplex are forward invariant (when started on one face, the trajectory of the replicator dynamics remains on that face). As well known, the game may not have an equilibrium on that face. Therefore, the outcome of the replicator dynamics may not be an equilibrium. To resolve this problem, one can fix the starting point at the relative interior of the simplex (for example, the uniform distribution can be chosen as initial point). Then, we have the following conclusions.
\begin{enumerate}[(S1)]
\item If started in the interior, the dominated strategies will be eliminated.
\item If started in the interior, and if the trajectory goes to the boundary, then the outcome is an equilibrium.
    \item If there is a cyclic orbit of the dynamics, the limit cycle contains an equilibrium in its interior.
        \item The expected payoff is learned if CODIPAS-RL CRL0 is used: $f(a_1)>0$ implies that $\hat{u}_{1,t}(a_1)\longrightarrow \mathbb{E}_{s} \mathbb{U}(s,e_{a_1},\mathbf{g}),$ and similarly for P2, $g(a_2)>0$ implies that $\hat{u}_{2,t}(a_2)\longrightarrow c-\mathbb{E}_{s} \mathbb{U}(s,\mathbf{f},e_{a_2}).$
\end{enumerate}
Another way of eliminating the non-equilibrium rest points is to perturb the game. The strategy can be perturbed using a small deviation from $(\mathbf{f},\mathbf{g}),$ i.e., an action $a_1$ will be chosen with probability $(1-\epsilon)f(a_1)+\frac{\epsilon}{|\mathcal{A}_1|}.$

{\it 2) Properties of CRL1 and CRL2: }
Numerically, the approximation of CRL0, RL2 and RL3 can lead to the boundary of the  simplex. To solve this problem, we propose a modified version of CODIPAS-RL based on Boltzmann-Gibbs distribution. These are the coupled reinforcement learning CRL1 and CRL2.
 Since the Boltzmann-Gibbs distribution never vanishes, the new algorithm CODIPAS-RL CRL1 based on Boltzmann-Gibbs is well defined for any initial condition and preserves the property that every rest point is a Boltzmann-Gibbs equilibrium, also called logit equilibrium, i.e., the fixed point of the mapping $\tilde{\beta}_{1,\epsilon}(\mathbb{E}_{s}\hat{u}_{1}({s},.,\mathbf{g}))=\mathbf{f}, \tilde{\beta}_{2,\epsilon}(\mathbb{E}_{s}\hat{u}_{2}({s},\mathbf{f},.))=\mathbf{g}$ which is an $\epsilon-$saddle-point equilibrium. Thus, by choosing $\epsilon$ arbitrarily small, an approximate solution is obtained. The main advantage of this Boltzmann-Gibbs distribution is that it is a smooth mapping (a regularized version of the best-response correspondence).
%\subsection{}
\section{Main results} \label{main}
In this section, we obtain ODE approximations of the learning algorithms in Section II and show the convergence of different heterogeneous learning algorithms to saddle-point solutions.

\subsection{Convergence to ODE: the combined learning algorithms}
We first examine the case where the players learn via different schemes but on the same time scale or by the same learning rate, i.e., the factor $\lambda_{i,t}=\lambda_t, i\in\{1, 2\},$ independent of the players. We use ${\beta}_{1,\epsilon}(\mathbf{g}_t): \Delta(\mathcal{A}_2)\rightarrow \Delta(\mathcal{A}_1)$ and ${\beta}_{2,\epsilon}(\mathbf{f}_t):\Delta(\mathcal{A}_1)\rightarrow \Delta(\mathcal{A}_2)$ to denote P1 and P2's Boltzmann-Gibbs responses to the other player's mixed strategies and ${\beta}_{1,\epsilon}(\mathbf{g}_t)(a_1):=\tilde{\beta}_{1,\epsilon}(u_1(e_{a_1},\mathbf{g}_t))$; ${\beta}_{2,\epsilon}(\mathbf{f}_t)(a_2):=\tilde{\beta}_{2,\epsilon}(u_2(\mathbf{f}_t, e_{a_2})), a_1\in\mathcal{A}_1, a_2\in\mathcal{A}_2.$
\medskip
\begin{subthm} \label{thmt1}
The combined learning algorithm with different learners using CRL1,  RL2,   RL3 converges to the joint system of ODEs. In particular, if P1 uses CRL1 and P2 adopts RL2, then the ODE is given by
\begin{equation}\label{EQNthm1}
\left\{\begin{array}{cll}
\frac{d}{dt}\hat{u}_{1,t}(a_1)&=&  {u}_{1}(e_{a_1},\mathbf{g}_t)- \hat{u}_{1,t}(a_1), \  a_1\in\mathcal{A}_1, \\
\dot{\mathbf{f}}_t&=& {\beta}_{1,\epsilon}(\mathbf{g}_t)-\mathbf{f}_t, \ \\
 \dot{g}_t(a_2)&=& {g}_t(a_2)[ u_2(\mathbf{f}_t,e_{a_2})\ \\ & & -\sum_{a'_2\in\mathcal{A}_2} u_2(\mathbf{f}_t,e_{a'_2}) g_t(a'_2) ], a_2\in\mathcal{A}_2.
\end{array}
\right.
\end{equation}
Moreover, if P2 adopts RL3 in lieu of RL2, then one has the {\it adjusted replicator dynamics} instead of the standard replicator equation.
\end{subthm}

We now have the following corollary corresponding to different learning rates for the two players.

\begin{subcoro} \label{corothm1} In the heterogeneous learning where players choose to adopt one learning scheme among CRL1, RL2, RL3 and with different learning rates, we have the following results.
\medskip

(C1)
If P1 uses CRL1 and P2 learns through RL2 with a rate $k_2$ faster than P1's rate, then the ODE is given by
{\small \begin{equation} \nonumber \left\{\begin{array}{ccl}
\frac{d}{dt}\hat{u}_{1,t}(a_1)&=&  {u}_{1}(e_{a_1},\mathbf{g}_t)- \hat{u}_{1,t}(a_1), \   a_1 \in \mathcal{A}_1 \\
\dot{\mathbf{f}}_t&=& {\beta}_{1,\epsilon}(\mathbf{g}_t)-\mathbf{f}_t  \ \\
 \dot{g}_t(a_2)&=& k_2 {g}_t(a_2)[ u_2(e_{a_2},\mathbf{f}_t),\ \\ & & -\sum_{a'_2\in\mathcal{A}_2} u_2(e_{a'_2},\mathbf{f}_t) g_t(a'_2) ] ,  a_2\in\mathcal{A}_2.
\end{array}
\right.
\end{equation}} Moreover, if P2 adopts RL3 in lieu of RL2, then one has the {\it $k_2-$adjusted replicator dynamics} instead of the standard replicator equation.
\medskip

(C2) If P1 uses CRL1 with a rate of learning $k_1$ faster than  P2 who learns with RL2, then the ODE is given by
{\small \begin{equation} \nonumber  \left\{\begin{array}{cll}
\frac{d}{dt}\hat{u}_{1,t}(a_1)&=&  {u}_{1}(e_{a_1},\mathbf{g}_t)- \hat{u}_{1,t}(a_1), \   a_1 \in \mathcal{A}_1,  \\
\dot{\mathbf{f}}_t&=& k_1\left[{\beta}_{1,\epsilon}(\mathbf{g}_t)-\mathbf{f}_t\right],  \ \\
 \dot{g}_t(a_2)&=& {g}_t(a_2)[ u_2(e_{a_2},\mathbf{f}_t)\ \\ & & -\sum_{a'_2\in\mathcal{A}_2} u_2(e_{a'_2},\mathbf{f}_t) g_t(a'_2)],  a_2 \in \mathcal{A}_2
\end{array}
\right.
\end{equation}}
\end{subcoro}

%\begin{proof} The proof follows the result of Theorem \ref{thmt1} by taking time scales $k_1\lambda_t,$ and $k_2\lambda_t$  in lieu of $ \lambda_t,$ and $\lambda_t$, respectively.
%\end{proof}

\begin{sublem}(\textbf{Explicit Solutions of Smooth BR Equation})\textbf{:} \label{lemt2}
Given P2's trajectory $\{\mathbf{g}_{t'}\}_{t'}$ and an initial condition $\mathbf{f}_0,$ the smooth best response equation
\begin{equation}\label{SBRE}
\dot{\mathbf{f}}_t={\beta}_{1,\epsilon}(\mathbf{g}_t)-\mathbf{f}_t
\end{equation}
 in (\ref{EQNthm1}) has a unique solution given by the vectorial function
\begin{equation}
\xi_1(\mathbf{g}_t)(a_1)={f}_0(a_1)e^{-t}+e^{-t}\int_0^t z_{1,t'}(a_1)\ e^{t'}dt' ,\ a_1\in\mathcal{A}_1,
\end{equation}
where $z_{1,t'}={\beta}_{1,\epsilon}(\mathbf{g}_{t'}).$
In particular, if P2 is a slow learner i.e., $\mathbf{g}_{t}=\mathbf{g},$ constant in time, then the smooth best response equation of P1 converges to
\begin{equation}
\xi_1(\mathbf{g})(a_1)=(1-e^{-t}){\beta}_{1,\epsilon}(\mathbf{g})(a_1)+e^{-t}f_0(a_1),\ a_1\in\mathcal{A}_1,
\end{equation}
 which goes to ${\beta}_{1,\epsilon}(\mathbf{g})$ when $t\longrightarrow +\infty.$
\end{sublem}

%\begin{proof}[Sketch of Proof]
% Let $\mathbf{g}$ be a stationary mixed strategy of P2. Then, the ODE of P1 when using smooth best response
%is given by
%\begin{eqnarray}
%\nonumber \dot{{f}}_{t}(a_1)&=&\frac{e^{\frac{u_1(e_{a_1},\mathbf{g})}{\epsilon}}}{\sum_{a_1'\in\mathcal{A}_1} e^{\frac{u_1(e_{a'_1},\mathbf{g})}{\epsilon}}}-{f}_{t}(a_1)\\
%\nonumber &:=&{\beta}_{1,\epsilon}(\mathbf{g})(a_1)-{f}_{t}(a_1),
%a_1\in\mathcal{A}_1.
%\end{eqnarray}
%We check   that the solution of this ODE is
%$$
%\mathbf{f}_t={\beta}_{1,\epsilon}(\mathbf{g})+(\mathbf{f}_{0}-{\beta}_{1,\epsilon}(\mathbf{g}))e^{-t}=\mathbf{f}_{0}e^{-t}+
%(1-e^{-t}){\beta}_{1,\epsilon}(\mathbf{g}),
%$$
%which is globally exponentially convergent to ${\beta}_{1,\epsilon}(\mathbf{g}).$
%%This completes the proof.
%\end{proof}
\begin{sublem}(\textbf{Explicit Solutions of Replicator Equation})\textbf{:} \label{lemt3}
Given P2's trajectory $\{\mathbf{g}_{t'}\}_{t'}$ and an interior initial condition $\mathbf{f}_0,$ the replicator equation in (\ref{EQNthm1}) has a unique solution given by the vectorial function
$
\xi_1(\mathbf{g}_t)(a_1)=\frac{e^{\int_0^t u_1(e_{a_1},\mathbf{g}_{t'})\ dt'}}{\sum_{a'_1\in\mathcal{A}_1}e^{\int_0^t u_1(e_{a'_1},\mathbf{g}_{t'})\ dt'}},\ a_1\in\mathcal{A}_1
$, with a normalization factor $f_0.$
In particular, if P2 is a slow learner, i.e. $\mathbf{g}_{t}=\mathbf{g}$, constant in time, then the replicator equation of  P1 converges to
$$
\xi_1(\mathbf{g})(a_1)=\frac{e^{t u_1(e_{a_1},\mathbf{g})}}{\sum_{a'_1\in\mathcal{A}_1}e^{t u_1(e_{a'_1},\mathbf{g})}},\ a_1\in\mathcal{A}_1.
$$
\end{sublem}
Note that these solutions are in the interior of the simplex for $t$ finite, but the trajectory can be arbitrarily close to the boundary when $t$ goes to infinity. In particular, if we assume that the other player is a slow learner, i.e., $\frac{\lambda_{2,t}}{\lambda_{1,t}}\to 0, $ then,
$$\xi_1(\mathbf{g})(a_1)(t) \rightarrow \frac{{f}_{0}(a_1)}{\sum_{a'_1\in BR_1(\mathbf{g})}\ {f}_{0}(a'_1) } \ind_{\{a_1\in BR_1(\mathbf{g})\}},$$ when $\epsilon\to 0.$ The set $BR_1(\mathbf{g})$ denotes the set of pure maximizers of $\mathbf{f}$ that  maximize $\mathbb{E}_s \mathbb{U}(s,\mathbf{f}, \mathbf{g}).$

%\begin{proof}[Sketch of Proof]  Given a trajectory $\{\mathbf{g}_t\},$ we verify that the function $\xi_1(\mathbf{g})(\cdot)$  satisfies  the replicator equation
%$$\dot{\mathbf{f}}_t(a_1)= {f}_t(a_1)\left[ u_1(e_{a_1},\mathbf{g}_{t})-\sum_{a'_1\in\mathcal{A}_1} u_1(e_{a'_1},\mathbf{g}_{t}){f}_t(a'_1)\right].$$ Fixing an initial condition $\mathbf{f}_0$ at the relative interior of the simplex, we use  Cauchy-Lipschitz theorem  to identify the unique solution at the interior of the simplex.
%
%\end{proof}
\begin{subpropo} \label{mainlimit}
Given any time-varying mixed strategies $\{\mathbf{g}_t\}_t,$ the explicit solution to the replicator equation is
$\xi_1(\mathbf{g}_t)(a_1)= \tilde{\beta}_{1,\frac{1}{t}}(V)(a_1)$, where $V$ is the payoff vector defined by $V(a_1):=u_1(e_{a_1},\bar{\mathbf{g}}_t)$, where $\bar{\mathbf{g}}_t=\frac{1}{t}\int_0^t \mathbf{g}_{t'}\ dt'.$ In particular,
 if  the time-average sequence $\bar{\mathbf{g}}_t$ converges to $ \bar{\mathbf{g}}_*,$ then the explicit solution $\xi_1(\mathbf{g}_t)$  converges to a smooth best response to $\bar{\mathbf{g}}_*.$
\end{subpropo}

%\begin{proof}
%Given any trajectory of mixed strategy $g_t$ of P2, the explicit solution
%is
%\begin{eqnarray}
%\nonumber \xi_1(\mathbf{g}_t)(a_1)&=&\frac{e^{\int_0^t u_1(e_{a_1},\mathbf{g}_{t'})\ dt'}}{\sum_{a'_1\in\mathcal{A}_1}e^{\int_0^t u_1(e_{a'_1},\mathbf{g}_{t'})\ dt'}}\\
%\nonumber &=&\frac{t V(a_1)}{\sum_{a'_1\in\mathcal{A}_1}e^{t V(a'_1)}}=
%\tilde{\beta}_{1,\frac{1}{t}}(V)(a_1),
%\end{eqnarray} where $V$ is the payoff vector defined by $$V(a_1)=\frac{1}{t}\int_0^t u_1(e_{a_1},\mathbf{g}_{t'})\ dt'= \frac{1}{t}\int_0^t \mathbb{E}_s \mathbb{U}(s,e_{a_1},\mathbf{g}_{t'})\ dt'.$$ Using the linearity of $u_1$ in the second argument, one gets
%\begin{eqnarray}
%\nonumber V(a_1)&=&\frac{1}{t}\int_0^t u_1(e_{a_1},\mathbf{g}_{t'})\ dt'\\
%\nonumber &=&u_1\left(e_{a_1},\frac{1}{t}\int_0^t \mathbf{g}_{t'}\ dt'\right)=u_1(e_{a_1},\bar{\mathbf{g}}_t),
%\end{eqnarray} where $\bar{\mathbf{g}}_t=\frac{1}{t}\int_0^t \mathbf{g}_{t'}\ dt'$ (the integral of vectorial function is taken component-wise). In particular, if  the time-average sequence $\bar{\mathbf{g}}_t$ converges to $ \bar{\mathbf{g}}_*,$ then the explicit solution $\xi_1(\mathbf{g}_t)=\tilde{\beta}_{1,\frac{1}{t}}(V)=
%\beta_{1,\frac{1}{t}}(\bar{\mathbf{g}}_t)$ converges to a smooth best response to $\bar{\mathbf{g}}_*.$ This completes the proof.
%\end{proof}

\begin{subthm}[Two Different Learners] \label{thmt2} Consider two learners: one learns faster than the other.

(T1) Assume that P1 is a slow learner of RL2 or RL3 and P2 is a fast learner of CRL1, i.e., $\frac{\lambda_{1,t}}{\lambda_{2,t}}\longrightarrow 0$ as $t\rightarrow \infty$ . Then almost surely,
    $
    \|\mathbf{g}_t-\xi_2(\mathbf{f})\|\longrightarrow 0
    $ as $t$ goes to infinity, where $\xi_2(\mathbf{f})={\beta}_{2,\epsilon}(\mathbf{f}),$ and
   {\small \begin{equation}\label{ODE1}
    \dot{f}_t(a_1)={f}_t(a_1)[u_1(e_{a_1},{\beta}_{2,\epsilon}(\mathbf{f}_t))-
    \sum_{a_1'\in\mathcal{A}_1}{f}_t(a'_1)u_1(e_{a_1'},{\beta}_{2,\epsilon}(\mathbf{f}_t))]
    \end{equation}} generates the asymptotic pseudo-trajectory of $\{\mathbf{f}_t\}_{t\geq 0}.$

(T2) Assume that P2 is slow learner of RL2 or RL3 and P1 is a fast learner of CRL1, i.e., $\frac{\lambda_{2,t}}{\lambda_{1,t}}\longrightarrow 0$ as $t\rightarrow \infty$ . Then, almost surely,
    $
    \|\mathbf{f}_t-\xi_1(\mathbf{g})\|\longrightarrow 0
    $ as $t$ goes to infinity, where
$$
\xi_1(\mathbf{g})(a_1)=\frac{e^{t u_1(e_{a_1},\mathbf{g})}}{\sum_{a'_1\in\mathcal{A}_1}e^{t u_1(e_{a'_1},\mathbf{g})}},\ a_1\in\mathcal{A}_1
$$
and the ODE
{ \begin{equation}\label{ODE2}
\dot{\mathbf{g}}_t={\beta}_{2,\epsilon}(\xi_1( \mathbf{g}_t))-\mathbf{g}_t
\end{equation}}
generates the asymptotic pseudo-trajectory of $\{\mathbf{g}_t\}_{t\geq 0}.$

\end{subthm}

Note that this last ODE differs from the replicator dynamics,
the   best response dynamics, the logit dynamics and
fictitious play, etc.

\begin{subrem}
Note that from Lemma \ref{lemt2}, $\xi_1(\mathbf{g})(a_1)=\beta_{1,\frac{1}{t}}(\mathbf{g})(a_1).$ This means that if the trajectories remain in the interior of the simplex, the time averages of the replicator dynamics and the smooth best-response dynamics are asymptotically close (the norm of the difference between the two trajectories is small when $t$ is sufficiently large). The mixed strategy $\beta_{1,\frac{1}{t}}$ has full support for any $t>0,$ i.e.,  $ \xi_1(\mathbf{g})$ remains in the relative interior of the simplex for all $t$.
\end{subrem}
The following theorem, whose proof can be found in the full report \cite{QTB}, says that under CRL1, the dominated strategies will be eliminated in the long-term.
\begin{subthm}
Consider algorithm CRL1. If a strategy $a_1$ is strictly dominated, then $f_{t}(a_1)\longrightarrow 0$ when $t \longrightarrow \infty$ and $\epsilon\longrightarrow 0.$
\end{subthm}

%\begin{proof}
%This follows from the fact that, given $\{\mathbf{g}_{t}\}_t$ the solution of the ODE
%$$\dot{f}_{t}(a_1)=\frac{e^{ \frac{u_1(e_{a_1},\mathbf{g}_{t})}{\epsilon}}}{\sum_{a'_1\in\mathcal{A}_1}e^{ \frac{u_1(e_{a'_1},\mathbf{g}_{t})}{\epsilon}}}-{f}_{t}(a_1),\ a_1\in\mathcal{A}_1,
%$$ is given by
%$$
%\xi_1(\mathbf{g}_{t})=f_{0}e^{-t}+e^{-t}\int_0^t {\beta}_{1,\epsilon}(\mathbf{g}_{t'})e^{t'}\ dt'.
%$$
%Since $\{\mathbf{g}_{t}\}_t$ is a sequence which takes value in a compact set (simplex), there
%exists a converging subsequence of $\mathbf{g}_{t}.$
%%Denote this subsequence by $y_{-j,\phi(t)}.$
%If $a_1$ is dominated, then $u_1(e_{a'_1},\mathbf{g}_{t})-u_1(e_{a_1},\mathbf{g}_{t})>\delta,\ \forall a'_1\neq a_1$ for some $\delta>0$ and hence
%\begin{eqnarray}
%\nonumber {\beta}_{1,\epsilon}(\mathbf{g}_{t})(a_1)&=&\frac{1}{1+\sum_{a'_1\neq a_1}e^{ \frac{u_1(e_{a'_1},\mathbf{g}_{t})-u_1(e_{a_1},\mathbf{g}_{t})}{\epsilon}}}\\
%\nonumber &\leq& \frac{1}{1+(|\mathcal{A}_1|-1)e^{ \frac{\delta}{\epsilon}}}
%\end{eqnarray}
%goes to zero because the denominator goes to infinity when $\epsilon\longrightarrow 0.$ Hence, we obtain
%$$\xi_1(\mathbf{g}_{t})(a_1)\leq f_{0}(a_1)e^{-t}+(1-e^{-t})\frac{1}{1+(|\mathcal{A}_1|-1)e^{ \frac{\delta}{\epsilon}}}\longrightarrow 0.$$
%\end{proof}

\subsection{Convergence to saddle points}
From (T1) of Theorem \ref{thmt2}, we see that the case with P1 as the slow learner leads to ODE in (\ref{ODE1}) whose solution is given by Lemma \ref{lemt3}, which is in the form of the smooth best response to P2. Knowing that $\mathbf{g}_t$ also converges almost surely to the smooth best response to P1, we conclude that the learning algorithm studied in (T1) converges to  an $\epsilon-$saddle point. Similarly, from (T2) of Theorem \ref{thmt2}, when P1 acts as a fast learner, the ODE in (\ref{ODE2}) has its solution given by Lemma \ref{lemt2} and leads to the smooth best response when $t\rightarrow \infty$. In addition,
from (T1) and from Proposition \ref{mainlimit}, $\mathbf{f}_t$ converges to $\xi_{1}=\beta_{1,\frac{1}{t}}$, which is asymptotically close to the smooth best-response dynamics. Hence we can conclude that the algorithm studied in (T2) also converges to an $\epsilon-$saddle point.
%we conclude that all the rest points of the heterogeneous dynamics are  smooth best reply points i.e., $\epsilon-$saddle  points. Thus,
When $\epsilon$ goes to zero, the stationary points of these heterogeneous dynamics converge to the saddle points of the expected game.
%We remark that $\xi_1(\mathbf{g})(a_1)=\beta_{1,\frac{1}{t}}(\mathbf{g})(a_1).$ This means that if the trajectories remains at the interior of the simplex, the time average of the replicator dynamics and the smooth best-response dynamics are asymptotically close. The mixed strategy $\beta_{1,\frac{1}{t}}$ has full support for any $t>0,$ i.e.,  $ \xi_1(\mathbf{g})$ remains at the relative interior of the simplex during the trajectory.
We can extend the preceding argument to any combination of replicator dynamics and smooth best response dynamics. Using Theorem \ref{thmt1} and its corollary \ref{corothm1}, we arrive at the following result.
 \begin{subthm}
 Consider the case of two different learners in which one learns faster than the other. Let the initial condition be an interior point of the simplex.
 The heterogeneous dynamics: (i) CRL0 with CRL1, (ii) CRL0 with CRL2, (iii)  CRL1 with CRL2, (iv) CRL1 with RL2, and (v) CRL1 with RL3 lead almost surely to an $\epsilon-$ saddle point of the expected game.
 \end{subthm}
 %The following result follows from \cite{hofbauer09,hofbauer98}.
%\begin{subthm}For the learning schemes RL2, RL3, CRL0,  the  time averages $\left(\frac{1}{T}\int_0^T \mathbf{f}_{t}(a_1)\ dt,\ \frac{1}{T}\int_0^T \mathbf{g}_{t}(a_2)\ dt\right)$ of an interior orbit converge to the set of saddle points.
%\end{subthm}
\begin{figure*}[t]
%\begin{center}
\begin{minipage}[b]{0.23\linewidth}
\centerline{\psfig{figure=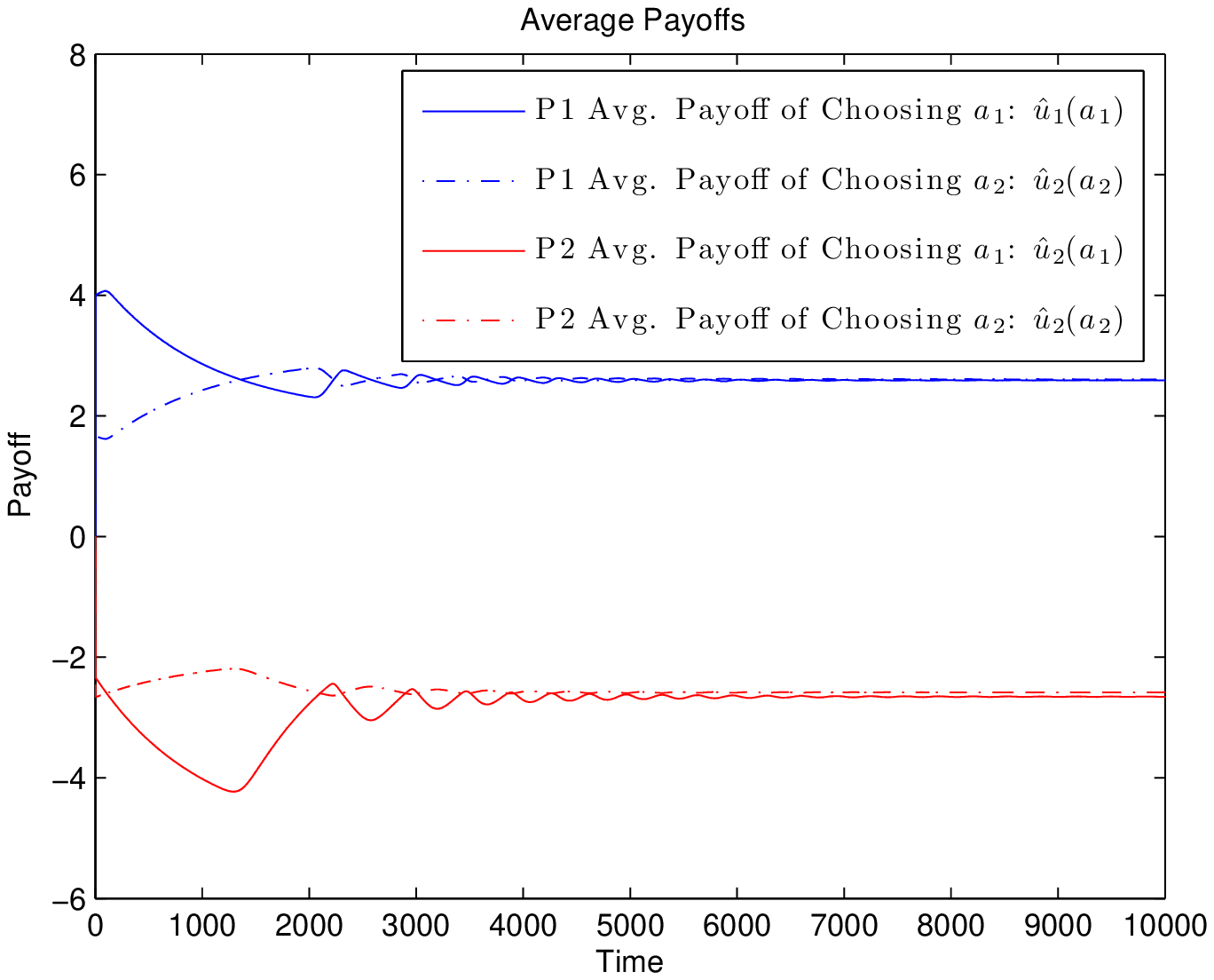,
scale=0.28}} \caption{The payoffs to the players with both players using CRL1.} \label{ODECRL1}
\end{minipage}
\hspace{0.3cm}
\begin{minipage}[b]{0.23\linewidth} % A minipage that covers half the page
\centerline{\psfig{figure=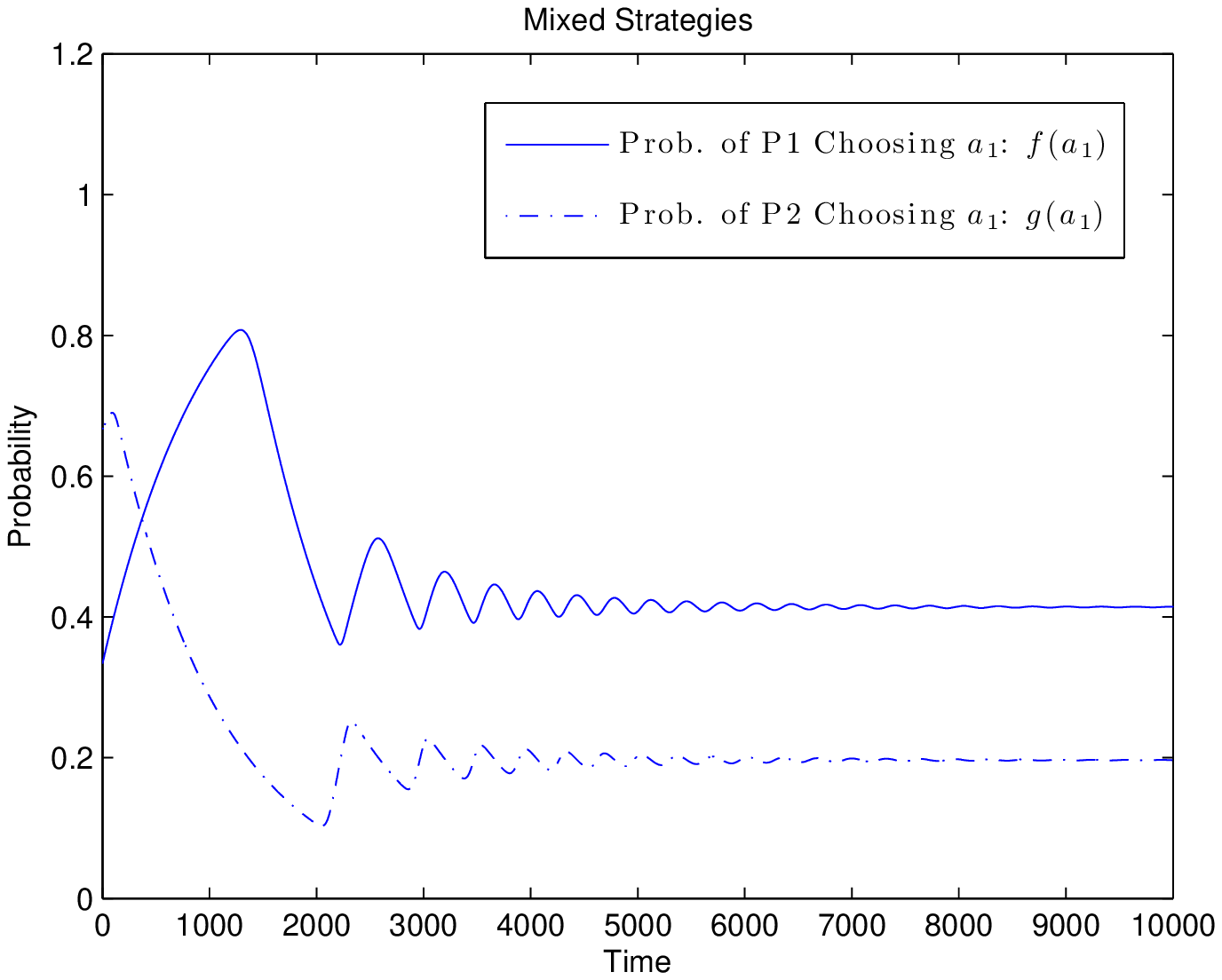,
scale= 0.28}} \caption{The mixed strategies of the players with both players using CRL1.} \label{ODECRL1MS}
\end{minipage}% \\
\hspace{0.3cm}
% To get a little bit of space between the figures
\begin{minipage}[b]{0.23\linewidth}
\centerline{\psfig{figure=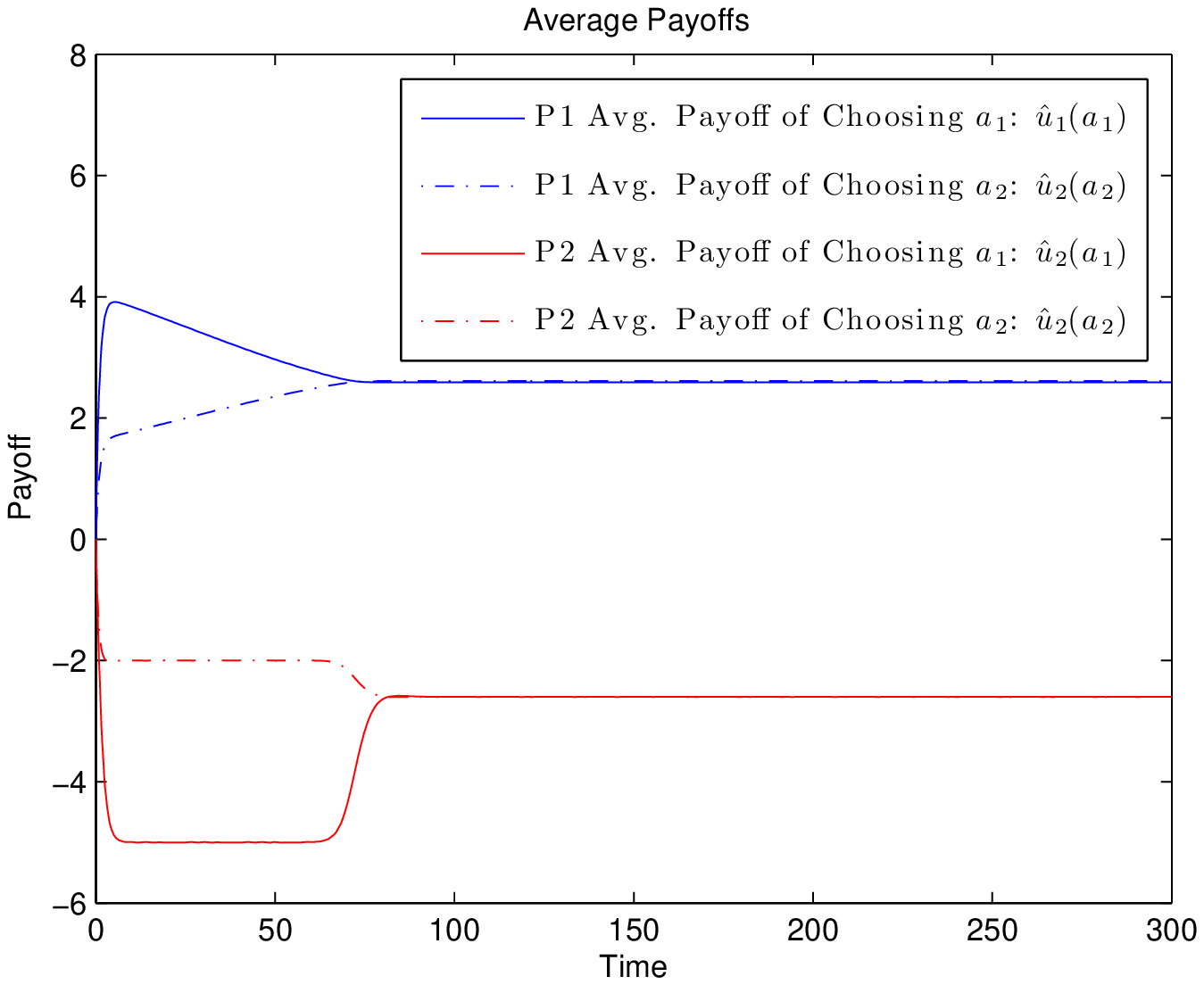,
scale= 0.28}} \caption{The payoffs to the players with the attacker using CRL1 and the defender using RL2.}
\label{ODECRL1RL2}
\end{minipage}
%\end{center}
%\end{figure*}
%
%\begin{figure*}[t]
%\begin{center}
\begin{minipage}[b]{0.23\linewidth}
\centerline{\psfig{figure= 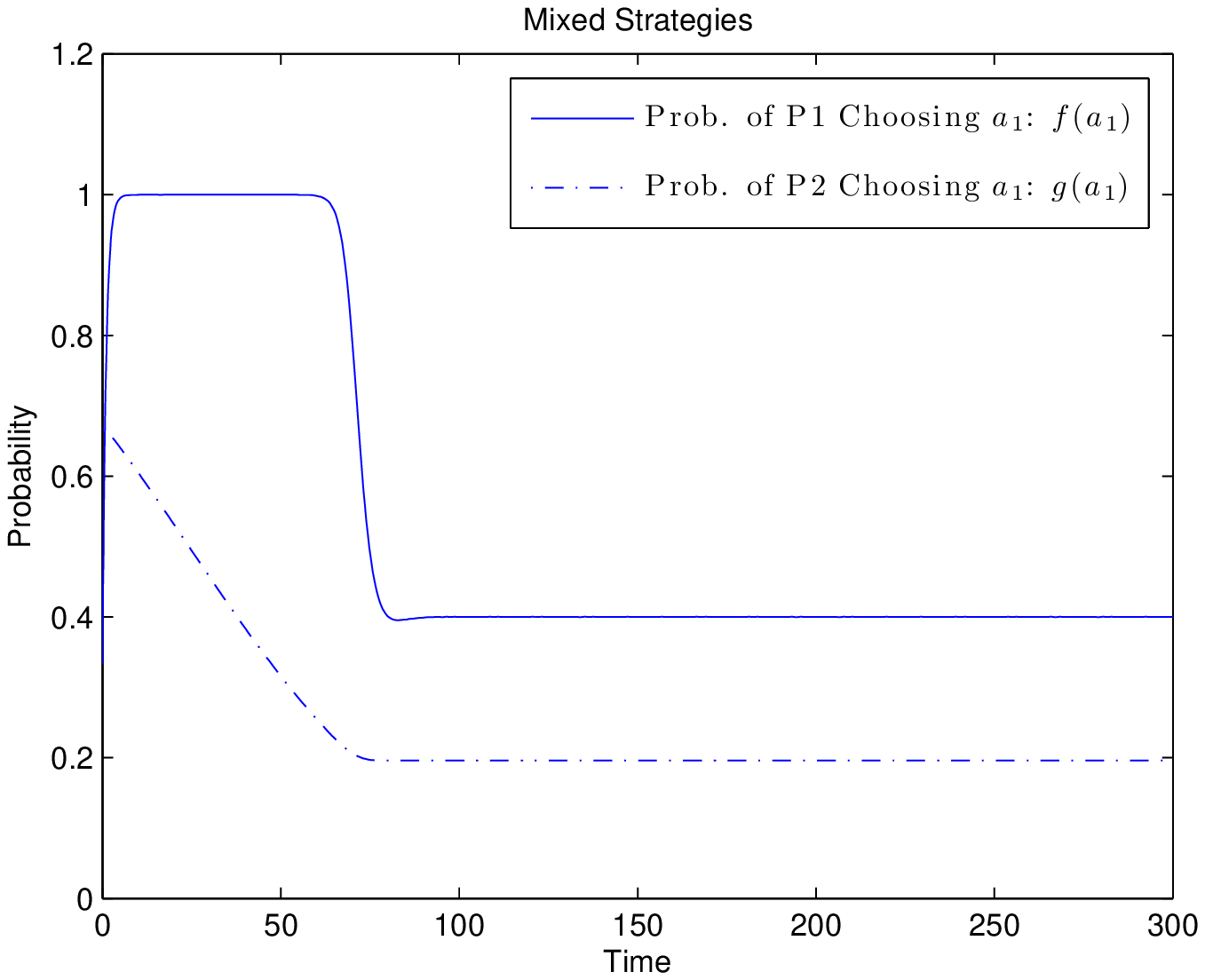,
scale= 0.28}} \caption{The mixed strategies of the players with the attacker using CRL1 and the defender using RL2.} \label{ODECRL1RL2MS}
\end{minipage}
%\vspace{-6mm}
\end{figure*}
%\vspace{-2mm}
\section{Application and Simulation}

In this section, we illustrate the heterogeneous learning algorithms with an example motivated by computer security.  In a network intrusion detection system, an intruder attempts to scan the host machines and seek their vulnerabilities while the intrusion detector monitors the suspicious behavior and raises an alarm when attacks are detected. The attacker and the defender can dynamically adapt their strategies from learning the history of the behaviors of each other and their own payoffs. It is common that the learning pattern of the attacker is different from the one used by the defender since learning schemes depend on an individual's preference and rationality as well as the information observed by each person. Hence, in the context of computer security,  heterogeneity of the learning algorithm is essential because it offers extra degrees of freedom to model agent's behavior.

Consider a two-person game with one party being the defender (P1) and the other party the attacker (P2).  The defender has two actions available for each play, i.e., either to defend (D) or not to defend (ND), while the attacker has two actions  either to attack or not to attack. The deterministic payoff matrix is given by
$ \mathbf{M}=\left[
\begin{array}{cc}
5 & 2 \\
1 & 3
\end{array}\right]
, $
where the columns correspond to the defender strategies (D) and (ND) whereas the rows correspond to the attacker strategies (A) and (NA). The stochastic payoff matrix $\mathbf{U}$ is a function of random matrix $\mathbf{S}=\left[
\begin{array}{cc}
s_1 & s_2 \\
s_3 & s_4
\end{array}\right]
, $ whose components are uniformly distributed on $[-1, 1]$.
It is given by
$ \mathbf{U}=\mathbf{M}+\mathbf{S}%=\left[
%\begin{array}{cc}
%5+s_1 & 2+s_2 \\
%1+ s_3 & 3+s_4
%\end{array}\right]
.$

At the equilibrium, the attacker selects its actions according to  $\mathbf{f}^*=[0.4, 0.6]^T$ while the defender chooses its actions using $\mathbf{g}^*=[0.2, 0.8]^T$. The strategy pair $(\mathbf{f}^*, \mathbf{g}^*)$ forms a saddle point solution to the game  $\mathbb{E}\mathbf{U}=\mathbf{M}$, yielding the game value $2.6$.
We show in Figures \ref{ODECRL1} and \ref{ODECRL1MS} the payoffs and the mixed strategies of the players, respectively, when both adopt the CRL1 learning algorithm. By setting $\epsilon=\frac{1}{20}$, we observe that the payoffs of P1 choosing actions N and NA at $t=8000$ are  $2.5890$ and  $2.6073$ respectively, which are close to the game value 2.6. For P2, the payoffs at $t=8000$ are $-2.6578$  and   $-2.5855$ for actions N and ND, respectively. The difference between the payoff and game value is explained by the soft-max parameter $\epsilon$. When $\epsilon$ approaches $0$, the average payoffs will approach the game value. The convergence of CRL1 is slow.  In Figures \ref{ODECRL1} and \ref{ODECRL1MS}, we observe that the payoff values and the mixed strategy probabilities converge roughly after $t=6000.$
In Figures \ref{ODECRL1RL2} and \ref{ODECRL1RL2MS}, we show the temporal evolution of the payoffs and mixed strategies of the attacker and defender using the heterogeneous learning algorithm in which the attacker follows CRL1 whereas the defender uses RL2. We initialize the payoffs to be $0$ and the strategy vectors $\mathbf{f}_0^T=[1/3, 2/3], \mathbf{g}_0^T=[1/3, 2/3]$. We set the parameter $\epsilon=\frac{1}{20}$ in the soft-max best response function of the attacker. The convergence of the learning process is shown after $t=80s$.
\section{Concluding remarks} \label{con}
 We have presented heterogeneous distributed
learning algorithms for two-person zero-sum stochastic games along with their general convergence and non-convergence
properties. Our results subsume many known
results regarding learning optimal strategies with different time scales and with different learning schemes. Interesting work that we leave for the future is to extend these results to stochastic games with controlled states and nonzero-sum stochastic games with incomplete information.
  \end{document}